\documentclass[journal]{IEEEtran}
\newcommand{\etal}{{\emph{et~al.}}~}

\usepackage{graphicx}
\usepackage{amsmath,amssymb}
\usepackage{bbm}
\usepackage{multirow}
\usepackage{caption}
\usepackage{algorithm}
\usepackage{algpseudocode}

\algdef{SE}[DOWHILE]{Do}{DoWhile}{\algorithmicdo}[1]{\algorithmicwhile\ #1}%
\def\NoNumber#1{{\def\alglinenumber##1{}\State #1}\addtocounter{ALG@line}{-1}}

\ifCLASSINFOpdf
\else
\fi
\begin{document}
%
\title{Skeleton Based Human Action Recognition with Global Context-Aware Attention LSTM Networks}
%
%
%

\author{Jun~Liu,~\IEEEmembership{Student Member,~IEEE,}
        Gang~Wang,~\IEEEmembership{Senior Member,~IEEE,}\\
        Ling-Yu~Duan,~\IEEEmembership{Member,~IEEE,}
        Kamila~Abdiyeva,~\IEEEmembership{Student Member,~IEEE,}
        and~Alex~C.~Kot,~\IEEEmembership{Fellow,~IEEE}
\thanks{J. Liu, K. Abdiyeva, and A. C. Kot are with School of Electrical and Electronic Engineering, Nanyang Technological University, Singapore, 639798. E-mail: \{jliu029, abdi0001, eackot\}@ntu.edu.sg.}
\thanks{G. Wang is with Alibaba Group, Hangzhou, 310052, China. Email: wanggang@ntu.edu.sg.}
\thanks{L.-Y. Duan is with National Engineering Lab for Video Technology, Peking University, Beijing, 100871, China. Email: lingyu@pku.edu.cn.}
\thanks{Manuscript received July 1, 2017.}}

%
%

\markboth{Accepted to IEEE Transactions on Image Processing}%
{Shell \MakeLowercase{\textit{et al.}}: Bare Demo of IEEEtran.cls for IEEE Journals}
%



\maketitle

\begin{abstract}
Human action recognition in 3D skeleton sequences has attracted a lot of research attention.
Recently, Long Short-Term Memory (LSTM) networks have shown promising performance in this task
due to their strengths in modeling the dependencies and dynamics in sequential data.
As not all skeletal joints are informative for action recognition,
and the irrelevant joints often bring noise which can degrade the performance,
we need to pay more attention to the informative ones.
However, the original LSTM network does not have explicit attention ability.
In this paper, we propose a new class of LSTM network, Global Context-Aware Attention LSTM (GCA-LSTM), for skeleton based action recognition,
which is capable of selectively focusing on the informative joints in each frame by using a global context memory cell.
To further improve the attention capability,
we also introduce a recurrent attention mechanism,
with which the attention performance of our network can be enhanced progressively.
Besides, a two-stream framework, which leverages coarse-grained attention and fine-grained attention, is also introduced.
The proposed method achieves state-of-the-art performance on five challenging datasets for skeleton based action recognition.
\end{abstract}


%
\IEEEpeerreviewmaketitle

%
%

\section{Introduction}
\label{sec:Introduction}

\begin{figure*}
	\centerline{\includegraphics[scale=0.32]{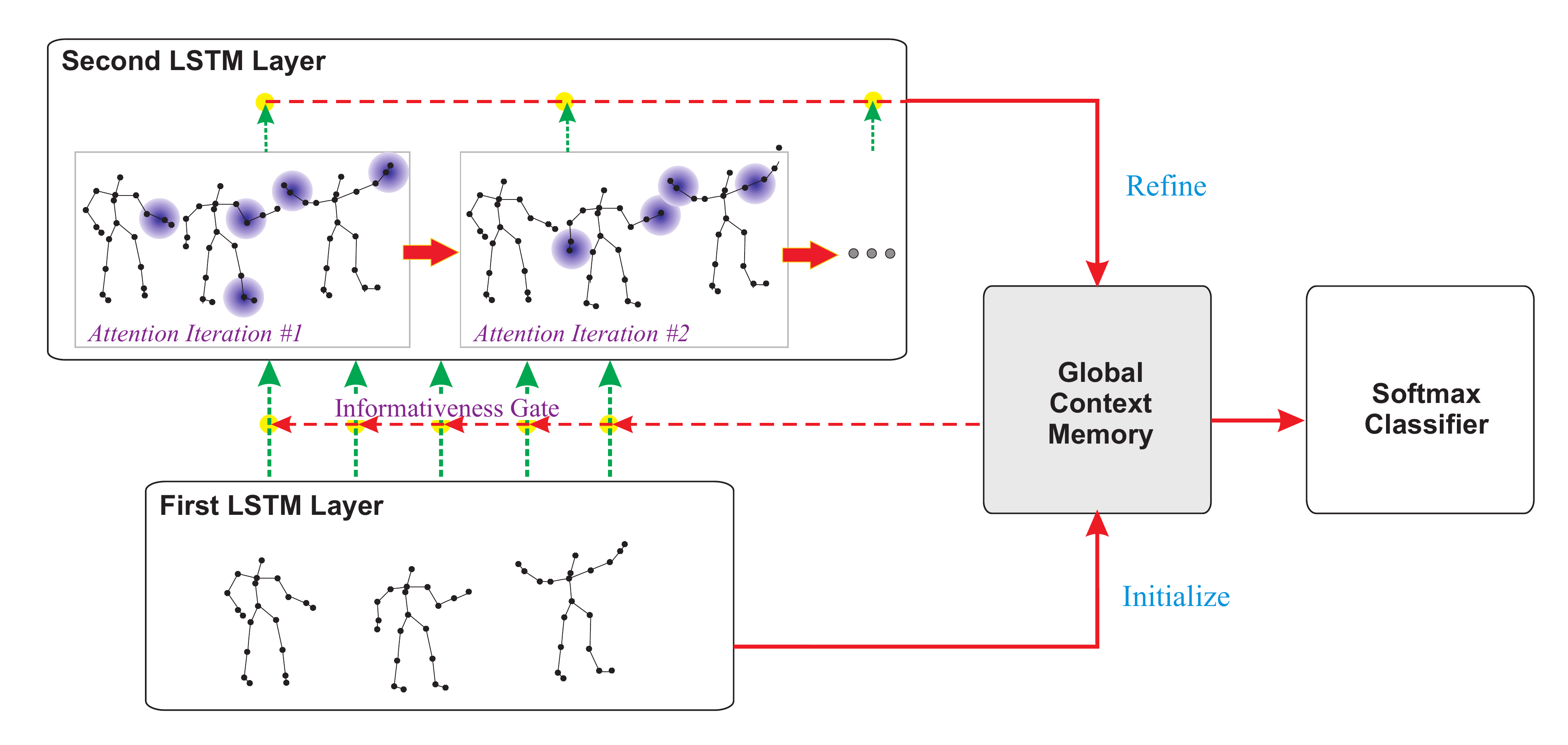}}
	\caption{Skeleton based human action recognition with the Global Context-Aware Attention LSTM network.
The first LSTM layer encodes the skeleton sequence and generates an initial global context representation for the action sequence.
The second layer performs attention over the inputs by using the global context memory cell to achieve an attention representation for the sequence.
Then the attention representation is used back to refine the global context.
Multiple attention iterations are performed to refine the global context memory progressively.
Finally, the refined global context information is utilized for classification.
}
	\label{fig:overview}
\end{figure*}

\IEEEPARstart{A}{ction} recognition is a very important research problem owing to its relevance to a wide range of applications,
such as video surveillance, patient monitoring, robotics, human-machine interaction, etc \cite{zheng2016cross,liu2016simple,jiang2015human}.
With the development of depth sensors, such as RealSense and Kinect \cite{han2013enhanced,zhang2017joint,zhang2016robust},
3D skeleton based human action recognition has received much attention,
and a lot of advanced methods have been proposed during the past few years \cite{presti20163d,han2016review,aggarwal2014human,zhang2016rgb}.

Human actions can be represented by a combination of the motions of skeletal joints in 3D space \cite{ye2013survey,du2015hierarchical}.
However, this does not indicate all joints in the skeleton sequence are informative for action recognition.
For instance, the hand joints' motions are quite informative for the action \emph{clapping},
while the movements of the foot joints are not.
Different action sequences often have different informative joints,
and in the same sequence, the informativeness degree of a body joint may also change over the frames.
Thus, it is beneficial to selectively focus on the informative joints in each frame of the sequence,
and try to ignore the features of the irrelevant ones,
as the latter contribute very little for action recognition,
and even bring noise which corrupts the performance \cite{jiang2015informative}.
This selectively focusing scheme can also be called \emph{attention},
which has been demonstrated to be quite useful for various tasks,
such as speech recognition \cite{chorowski2015attention}, image caption generation \cite{xu2015show}, machine translation \cite{bahdanau2014neural}, and so on.

Long Short-Term Memory (LSTM) networks have strong power in handling sequential data \cite{hochreiter1997long}.
They have been successfully applied to language modeling \cite{sundermeyer2012lstm},
RGB based video analysis \cite{ibrahim2016hierarchical,yeung2016end,yue2015beyond,wu2015modeling,donahue2015long,keleveraging,ke2016spatial,srivastava2015unsupervised,ma2016learning},
and also skeleton based action recognition \cite{du2015hierarchical,zhu2016co,liu2016eccv}.
However, the original LSTM does not have strong attention capability for action recognition.
This limitation is mainly owing to LSTM's restriction in perceiving the global context information of the video sequence, which is, however, often very important for the global classification problem \--- skeleton based action recognition.

In order to perform reliable attention over the skeletal joints, we need to assess the informativeness degree of each joint in each frame with regarding to the global action sequence.
This indicates that we need to have global contextual knowledge first.
However, the available context information at each evolution step of LSTM is relatively local.
In LSTM, the sequential data is fed to the network as input step by step.
Accordingly, the context information (hidden representation) of each step is fed to the next one.
This implies the available context at each step is the hidden representation from the previous step, which is quite local when compared to the global information
\footnote{Though in LSTM, the hidden representations of the latter steps contain wider range of context information than that of the initial steps, their context is still relatively local, as LSTM has trouble in remembering information too far in the past \cite{weston2015memory}. }.

In this paper, we extend the original LSTM model
and propose a Global Context-Aware Attention LSTM (GCA-LSTM) network which has strong attention capability for skeleton based action recognition.
In our method, the global context information is fed to all evolution steps of the GCA-LSTM.
Therefore, the network can use it to measure the informativeness scores of the new inputs at all steps,
and adjust the attention weights for them accordingly,
i.e., if a new input is informative regarding to the global action, then the network takes advantage of more information of it at this step,
on the contrary, if it is irrelevant, then the network blocks the input at this step.

Our proposed GCA-LSTM network for skeleton based action recognition includes a global context memory cell and two LSTM layers, as illustrated in \figurename{~\ref{fig:overview}}.
The first LSTM layer is used to encode the skeleton sequence and initialize the global context memory cell.
And the representation of the global context memory is then fed to the second LSTM layer to
assist the network to selectively focus on the informative joints in each frame, and further generate an attention representation for the action sequence.
Then the attention representation is fed back to the global context memory cell in order to refine it.
Moreover, we propose a recurrent attention mechanism for our GCA-LSTM network.
As a refined global context memory is produced after the attention procedure,
the global context memory can be fed to the second LSTM layer again to perform attention more reliably.
We carry out multiple attention iterations to optimize the global context memory progressively.
Finally, the refined global context is fed to the softmax classifier to predict the action class.

In addition, we also extend the aforementioned design of our GCA-LSTM network in this paper, and further propose a two-stream GCA-LSTM,
which incorporates fine-grained (joint-level) attention and coarse-grained (body part-level) attention,
in order to achieve more accurate action recognition results.

The contributions of this paper are summarized as follows:
\begin{itemize}
   \item A GCA-LSTM model is proposed,
   which retains the sequential modeling ability of the original LSTM,
   meanwhile promoting its selective attention capability by introducing a global context memory cell.
   \item A recurrent attention mechanism is proposed, with which the attention performance of our network can be improved progressively.
   \item A stepwise training scheme is proposed to more effectively train the network.
   \item We further extend the design of our GCA-LSTM model, and propose a more powerful two-stream GCA-LSTM network.
   \item The proposed end-to-end network yields state-of-the-art performance on the evaluated benchmark datasets.
\end{itemize}

This work is an extension of our preliminary conference paper \cite{liu2017global}.
Based on the previous version, we further propose a stepwise training scheme to train our network effectively and efficiently.
Moreover, we extend our GCA-LSTM model and further propose a two-stream GCA-LSTM by leveraging fine-grained attention and coarse-grained attention.
Besides, we extensively evaluate our method on more benchmark datasets.
More empirical analysis of the proposed approach is also provided.

The rest of this paper is organized as follows.
In Section \ref{sec:related_work}, we review the related works on skeleton based action recognition.
In Section \ref{sec:approach}, we introduce the proposed GCA-LSTM network. 
In Section \ref{sec:approach2stream}, we introduce the two-stream attention framework. 
We provide the experimental results in Section \ref{sec:experiments}.
Finally, we conclude the paper in Section \ref{sec:conclusion}.

\section{Related Work}
\label{sec:related_work}

In this section, we first briefly review the skeleton based action recognition methods which mainly focus on extracting hand-crafted features.
We then introduce the RNN and LSTM based methods.
Finally, we review the recent works on attention mechanism.

\subsection{Skeleton Based Action Recognition with Hand-crafted Features}
In the past few years, different feature extractors and classifier learning methods for skeleton based action recognition have been proposed
\cite{luo2013group,
MMMP_PAMI,
meng_interaction_FG,
yang2014effective,
MMTW_iccv13,
Lillo_2016_CVPR,
rahmani2014real,
chen_2016_icassp,
Tao_2015_ICCV_Workshops,
shahroudy2014multi,
wang2014mining,
ofli2014sequence,
anirudh2015elastic}.

Chaudhry \etal \cite{chaudhry2013bio} proposed to encode the skeleton sequences to spatial-temporal hierarchical models,
and then use linear dynamical systems (LDSs) to learn the dynamic structures.
Vemulapalli \etal \cite{vemulapalli2014liegroup} represented each action as a curve in a Lie group,
and then utlized a support vector machine (SVM) to classify the actions.
Xia \etal \cite{xia2012view} proposed to model the temporal dynamics in action sequences with the Hidden Markov models (HMMs).
Wang \etal \cite{wang2012mining,wang2014learning} introduced an actionlet ensemble representation to model the actions meanwhile capturing the intra-class variances.
Chen \etal \cite{chen2016novel} designed a part-based 5D feature vector to explore the relevant joints of body parts in skeleton sequences.
Koniusz \etal \cite{koniusz2016tensor} introduced tensor representations for capturing the high-order relationships among body joints.
Wang \etal \cite{wang2016graph} proposed a graph-based motion representation in conjunction with a SPGK-kernel SVM for skeleton based activity recognition.
Zanfir \etal \cite{zanfir2013moving} developed a moving pose framework together with a modified k-NN classifier for low-latency action recognition.

\subsection{Skeleton Based Action Recognition with RNN and LSTM Models}
Very recently, deep learning, especially recurrent neural network (RNN), based approaches have shown their strength in skeleton based action recognition.
Our proposed GCA-LSTM network is based on the LSTM model which is an extension of RNN.
In this part, we review the RNN and LSTM based methods as below, since they are relevant to our method.

Du \etal \cite{du2015hierarchical} introduced a hierarchical RNN model to represent the human body structure and temporal dynamics of the joints.
Veeriah \etal \cite{veeriah2015differential} proposed a differential gating scheme to make the LSTM network emphasize on the change of information.
Zhu \etal \cite{zhu2016co} proposed a mixed-norm regularization method for the LSTM network in order to drive the model towards learning co-occurrence features of the skeletal joints.
They also designed an in-depth dropout mechanism to effectively train the network.
Shahroudy \etal \cite{Shahroudy_2016_CVPR} introduced a part-aware LSTM model to push the network towards learning long-term context representations of different body parts separately.
Liu \etal \cite{liu2016eccv,liu2017skeleton} designed a 2D Spatio-Temporal LSTM framework to concurrently explore the hidden sources of action related context information in both temporal and spatial domains.
They also introduced a trust gate mechanism \cite{liu2016eccv} to deal with the inaccurate 3D coordinates of skeletal joints provided by the depth sensors.

Beside action recognition,
RNN and LSTM models have also been applied to skeleton based action forecasting \cite{jain2015structural} and detection \cite{li2016online,jain2015structural}.

Different from the aforementioned RNN/LSTM based approaches, 
which do not explicitly consider the informativeness of each skeletal joint with regarding to the global action sequence,
our proposed GCA-LSTM network utilizes the global context information to perform attention over all the evolution steps of LSTM to selectively emphasize the informative joints in each frame,
and thereby generates an attention representation for the sequence, which can be used to improve the classification performance.
Furthermore, a recurrent attention mechanism is proposed to iteratively optimize the attention performance.

\subsection{Attention Mechanism}
Our approach is also related to the attention mechanism \cite{chorowski2015attention,bahdanau2014neural,xiong2016dynamic,sharma2015action,kumar2016ask,luong2015effective,sukhbaatar2015end}
which allows the networks to selectively focus on specific information.
Luong \etal \cite{luong2015effective} proposed a network with attention mechanism for neural machine translation.
Stollenga \etal \cite{stollenga2014deep} designed a deep attention selective network for image classification.
Xu \etal \cite{xu2015show} proposed to incorporate hard attention and soft attention for image caption generation.
Yao \etal \cite{yao2015describing} introduced a temporal attention model for video caption generation.

Though a series of deep learning based models have been proposed for video analysis in existing works
\cite{simonyan2014two,
shahroudy2016deep},
most of them did not consider the attention mechanism.
There are several works which explored attention, such as the methods in \cite{sharma2015action,song2017end,wang2016hierarchical}.
However, our method is significantly different from them in the following aspects:
These works use the hidden state of the previous time step of LSTM, whose context information is quite local, to measure the attention scores for the next time step.
For the global classification problem - action recognition,
the global information is crucial for reliably evaluating the importance (informativeness) of each input to achieve a reliable attention.
Therefore, we propose a global context memory cell for LSTM,
which is utilized to measure the informativeness score of the input at each step.
Then the informativeness score is used as a gate (informativeness gate, similar to the input gate and forget gate) inside the LSTM unit to adjust the contribution of the input data at each step for updating the memory cell.
To the best of our knowledge, we are the first to introduce a global memory cell for LSTM network to handle global classification problems.
Moreover, a recurrent attention mechanism is proposed to iteratively promote the attention capability of our network,
while the methods in \cite{sharma2015action,song2017end,wang2016hierarchical} performed attention only once.
In addition, a two-stream attention framework incorporating fine-grained attention and coarse-grained attention is also introduced.
Owing to the new contributions, our proposed network yields state-of-the-art performance on the evaluated benchmark datasets.

\section{GCA-LSTM Network}
\label{sec:approach}

In this section, we first briefly review the 2D Spatio-Temporal LSTM (ST-LSTM) as our base network.
We then introduce our proposed Global Context-Aware Attention LSTM (GCA-LSTM) network in detail,
which is able to selectively focus on the informative joints in each frame of the skeleton sequence by using global context information.
Finally, we describe our approach to training our network effectively.

\subsection{Spatio-Temporal LSTM}
\label{sec:approach:stlstm}

In a generic skeleton based human action recognition problem, the 3D coordinates of the major body joints in each frame are provided.
The spatial dependence of different joints in the same frame and the temporal dependence of the same joint among different frames are both crucial cues for skeleton based action analysis.
Very recently, Liu \etal \cite{liu2016eccv} proposed a 2D ST-LSTM network for skeleton based action recognition,
which is capable of modeling the dependency structure and context information in both spatial and temporal domains simultaneously.

\begin{figure}
	\centerline{\includegraphics[scale=0.50]{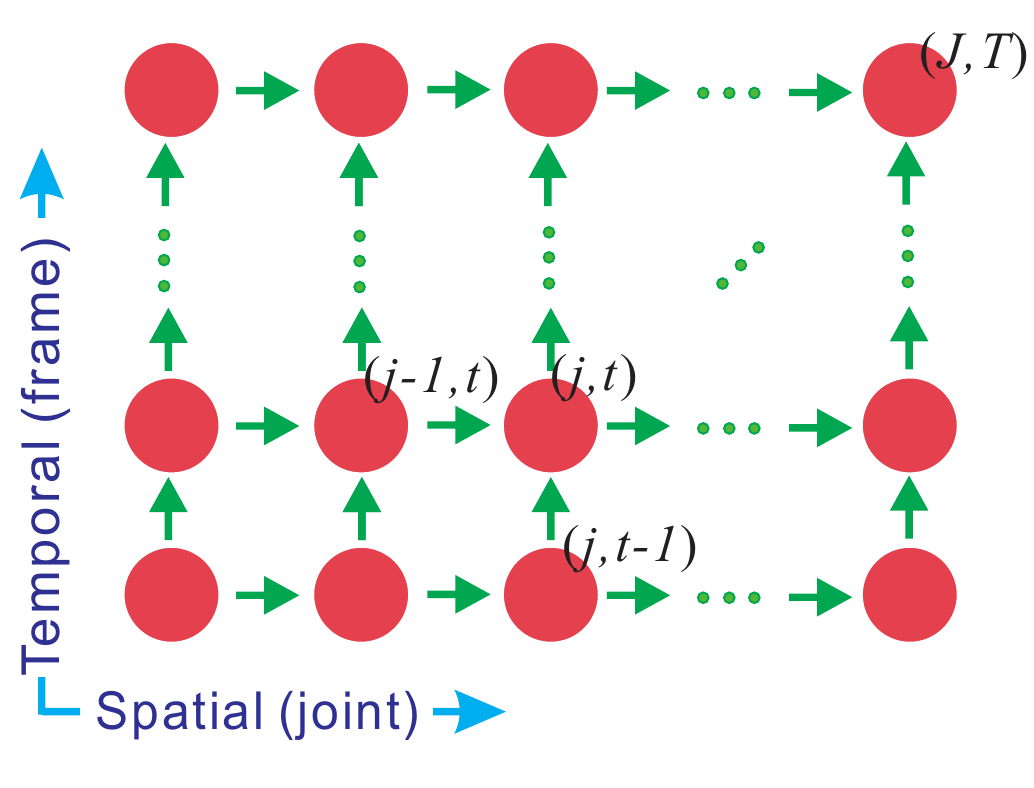}}
	\caption{Illustration of the ST-LSTM network \cite{liu2016eccv}.
    In the spatial direction, the body joints in each frame are arranged as a chain and fed to the network as a sequence.
    In the temporal dimension, the body joints are fed over the frames.
    }
	\label{fig:STLSTM}
\end{figure}

As depicted in \figurename{ \ref{fig:STLSTM}},
in ST-LSTM model, the skeletal joints in a frame are arranged and fed as a chain (the spatial direction),
and the corresponding joints over different frames are also fed in a sequence (the temporal direction).

Specifically, each ST-LSTM unit is fed with a new input ($x_{j, t}$, the 3D location of joint $j$ in frame $t$),
the hidden representation of the same joint at the previous time step $(h_{j,t-1})$,
and also the hidden representation of the previous joint in the same frame $(h_{j-1,t})$,
where $j \in \{1,...,J\}$ and $t \in \{1,...,T\}$ denote the indices of joints and frames, respectively.
The ST-LSTM unit has an input gate $(i_{j, t})$,
two forget gates corresponding to the two sources of context information ($f_{j, t}^{(T)}$ for the temporal dimension, and $f_{j, t}^{(S)}$ for the spatial domain),
together with an output gate $(o_{j, t})$.

The transition equations of ST-LSTM are formulated as presented in \cite{liu2016eccv}:
\begin{eqnarray}
\left(
   \begin{array}{ccc}
    i_{j, t} \\
    f_{j, t}^{(S)} \\
    f_{j, t}^{(T)} \\
    o_{j, t} \\
    u_{j, t} \\
   \end{array}
\right)
&=&
\left(
   \begin{array}{ccc}
    \sigma \\
    \sigma \\
    \sigma \\
    \sigma \\
    \tanh \\
   \end{array}
\right)
\left(
   W
   \left(
       \begin{array}{ccc}
        x_{j, t} \\
        h_{j-1, t} \\
        h_{j, t-1} \\
       \end{array}
   \right)
\right)
\\\nonumber
\label{eq:c_j_t_original}
c_{j, t} &=&  i_{j, t} \odot u_{j, t} \\ && + ~ f_{j, t}^{(S)} \odot  c_{j-1, t} \\\nonumber && + ~ f_{j, t}^{(T)} \odot  c_{j, t-1}
\\
\label{eq:h_j_t_original}
h_{j, t} &=& o_{j, t}  \odot \tanh( c_{j, t})
\end{eqnarray}
where $c_{j, t}$ and $h_{j, t}$ denote the cell state and hidden representation of the unit at the spatio-temporal step $(j, t)$, respectively,
$u_{j, t}$ is the modulated input, $\odot$ denotes the element-wise product,
and $W$ is an affine transformation consisting of model parameters.
Readers are referred to \cite{liu2016eccv} for more details about the mechanism of ST-LSTM.

\subsection{Global Context-Aware Attention LSTM}
\label{sec:approach:gcalstm}

\begin{figure}
	\centerline{\includegraphics[scale=0.369]{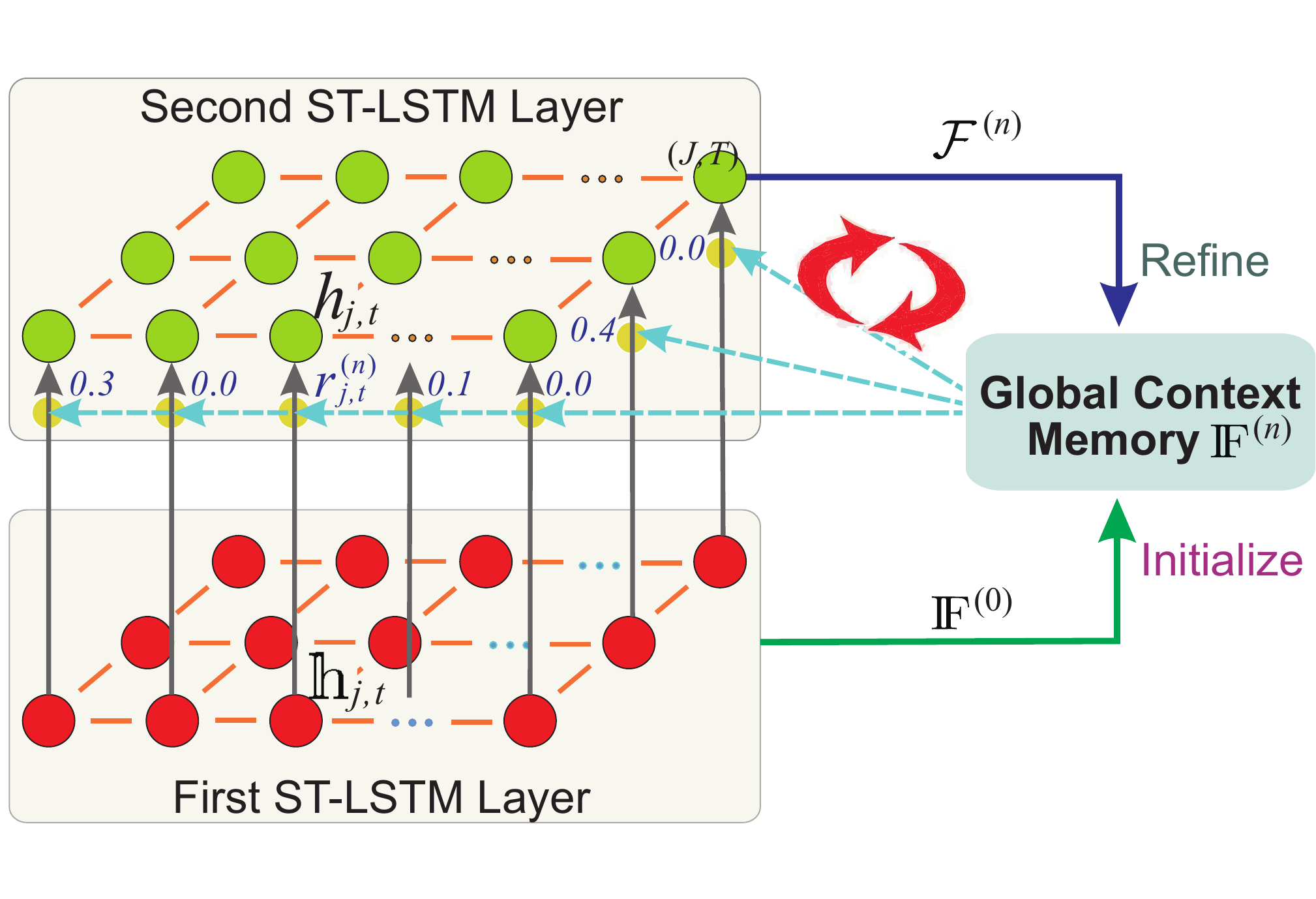}}
	\caption{Illustration of our GCA-LSTM network. Some arrows are omitted for clarity.}
	\label{fig:AttentionGateSTLSTM}
\end{figure}

Several previous works \cite{jiang2015informative,chen2016novel} have shown that in each action sequence,
there is often a subset of informative joints which are important as they contribute much more to action analysis,
while the remaining ones may be irrelevant (or even noisy) for this action.
As a result, to obtain a high accuracy of action recognition,
we need to identify the informative skeletal joints and concentrate more on their features,
meanwhile trying to ignore the features of the irrelevant ones,
i.e., selectively focusing (\emph{attention}) on the informative joints is useful for human action recognition.

Human action can be represented by a combination of skeletal joints' movements.
In order to reliably identify the informative joints in an action instance,
we can evaluate the informativeness score of each joint in each frame with regarding to the global action sequence.
To achieve this purpose, we need to obtain the global context information first.
However, the available context at each evolution step of LSTM is the hidden representation from the previous step,
which is relatively local when compared to the global action.

To mitigate the aforementioned limitation,
we propose to introduce a global context memory cell for the LSTM model,
which keeps the global context information of the action sequence, and can be fed to each step of LSTM to assist the attention procedure,
as illustrated in \figurename{~\ref{fig:AttentionGateSTLSTM}}.
We call this new LSTM architecture as Global Context-Aware Attention LSTM (GCA-LSTM).

\subsubsection{Overview of the GCA-LSTM network}

We illustrate the proposed GCA-LSTM network for skeleton based action recognition in \figurename{~\ref{fig:AttentionGateSTLSTM}}.
Our GCA-LSTM network contains three major modules.
The \emph{global context memory cell} maintains an overall representation of the whole action sequence.
The \emph{first ST-LSTM layer} encodes the skeleton sequence, and initializes the global context memory cell.
The \emph{second ST-LSTM layer} performs attention over the inputs at all spatio-temporal steps to generate an attention representation of the action sequence,
which is then used to refine the global context memory.

The input at the spatio-temporal step $(j,t)$ of the first ST-LSTM layer is the 3D coordinates of the joint $j$ in frame $t$.
The inputs of the second layer are the hidden representations from the first layer.

Multiple attention iterations (recurrent attention) are performed in our network to refine the global context memory iteratively.
Finally, the refined global context memory can be used for classification.

To facilitate our explanation, we use $\mathbbm{h}_{j, t}$
instead of $h_{j, t}$ to denote the hidden representation at the step $(j,t)$ in the first ST-LSTM layer,
while the symbols, including $h_{j, t}$, $c_{j, t}$, $i_{j, t}$,  and $o_{j, t}$,
which are defined in Section \ref{sec:approach:stlstm}, are utilized to represent the components in the second layer only.

\subsubsection{Initializing the Global Context Memory Cell}
\label{sec:approach:init}

Our GCA-LSTM network performs attention by using the global context information,
therefore, we need to obtain an initial global context memory first.

A feasible scheme is utilizing the outputs of the first layer to generate a global context representation.
We can average the hidden representations at all spatio-temporal steps of the first layer to compute an initial global context memory cell (${\rm I\!F}^{(0)}$) as follows:
\begin{eqnarray}
\label{eq:initialglobalcontext}
{\rm I\!F}^{(0)} = \frac{1}{JT}\sum\limits_{j=1}^J \sum\limits_{t=1}^T  \mathbbm{h}_{j,t}
\end{eqnarray}

We may also concatenate the hidden representations of the first layer and feed them to a feed-forward neural network, then use the resultant activation as ${\rm I\!F}^{(0)}$.
We empirically observe these two initialization schemes perform similarly.

\subsubsection{Performing Attention in the Second ST-LSTM Layer}
By using the global context information, we evaluate the informativeness degree of the input at each spatio-temporal step in the second ST-LSTM layer.

In the $n$-th attention iteration, our network learns an informativeness score ($r_{j,t}^{(n)}$) for each input ($\mathbbm{h}_{j,t}$) by
feeding the input itself, together with the global context memory cell (${\rm I\!F}^{(n-1)}$) generated by the previous attention iteration to a network as follows:
\begin{equation}
\label{eq:e_j_t}
e_{j,t}^{(n)} = W_{e_1}
    \left(
        \tanh
        \left(
            W_{e_2}
            \left(
                \begin{array}{ccc}
                \mathbbm{h}_{j,t} \\
                {\rm I\!F}^{(n-1)} \\
                \end{array}
            \right)
        \right)
   \right)
\end{equation}
\begin{equation}
\label{eq:r_j_t}
r_{j,t}^{(n)} = \frac{\exp ( e_{j,t}^{(n)} )}{ \sum\limits_{u=1}^J \sum\limits_{v=1}^T \exp ( {e_{u,v}^{(n)}} ) }
\end{equation}
where $r_{j,t}^{(n)} \in (0,1)$ denotes the normalized informativeness score of the input at the step $(j,t)$ in the $n$-th attention iteration,
with regarding to the global context information.

The informativeness score $r_{j,t}^{(n)}$ is then used as a gate of the ST-LSTM unit, and we call it \emph{informativeness gate}.
With the assistance of the learned informativeness gate, the cell state of the unit in the second ST-LSTM layer can be updated as:
\begin{eqnarray}
\label{eq:gatingstlstm_c}
\nonumber
c_{j, t} &=&          r_{j, t}^{(n)}  \odot i_{j, t} \odot u_{j, t} \\
          && + ~ (1 - r_{j, t}^{(n)}) \odot f_{j, t}^{(S)} \odot  c_{j-1, t} \\\nonumber
          && + ~ (1 - r_{j, t}^{(n)}) \odot f_{j, t}^{(T)} \odot  c_{j, t-1}
\end{eqnarray}

The cell state updating scheme in Eq. (\ref{eq:gatingstlstm_c}) can be explained as follows:
(1) if the input $(\mathbbm{h}_{j,t})$ is informative (important) with regarding to the global context representation,
then we let the learning algorithm update the cell state of the second ST-LSTM layer by importing more information of it;
(2) on the contrary, if the input is irrelevant, then we need to block the input gate at this step, meanwhile relying more on the history information of the cell state.

\subsubsection{Refining the Global Context Memory Cell}
We perform attention by adopting the cell state updating scheme in Eq. (\ref{eq:gatingstlstm_c}),
and thereby obtain an attention representation of the action sequence.
Concretely, the output of the last spatio-temporal step in the second layer is used as the attention representation ($\mathcal{F}^{(n)}$) for the action.
Finally, the attention representation $\mathcal{F}^{(n)}$ is fed to the global context memory cell to refine it, as illustrated in \figurename{~\ref{fig:AttentionGateSTLSTM}}.
The refinement is formulated as follows:
\begin{eqnarray}
\label{eq:updatingrnn}
{\rm I\!F}^{(n)} = {\rm ReLu}
    \left(
        W_{F}^{(n)}
        \left(
                   \begin{array}{ccc}
                    \mathcal{F}^{(n)}\\
                    {\rm I\!F}^{(n-1)} \\
                   \end{array}
        \right)
   \right)
\end{eqnarray}
where ${\rm I\!F}^{(n)}$ is the refined version of ${\rm I\!F}^{(n-1)}$.
Note that $W_{F}^{(n)}$ is not shared over different iterations.

Multiple attention iterations (recurrent attention) are carried out in our GCA-LSTM network.
Our motivation is that after we obtain a refined global context memory cell,
we can use it to perform the attention again to more reliably identify the informative joints, and thus achieve a better attention representation,
which can then be utilized to further refine the global context.
After multiple iterations, the global context can be more discriminative for action classification.

\subsubsection{Classifier}
The last refined global context memory cell ${\rm I\!F}^{(N)}$ is fed to a softmax classifier to predict the class label:
\begin{eqnarray}
\hat{y} = {\rm softmax}\left(W_c \left({\rm I\!F}^{(N)} \right)\right)
\end{eqnarray}

The negative log-likelihood loss function \cite{graves2012supervised} is adopted to measure the difference between the true class label $y$ and the prediction result $\hat{y}$.
The back-propagation algorithm is used to minimize the loss function.
The details of the back-propagation process are described in Section \ref{sec:approach:training}.

%

\subsection{Training the Network}
\label{sec:approach:training}

\begin{figure*}
	\centerline{\includegraphics[scale=0.28]{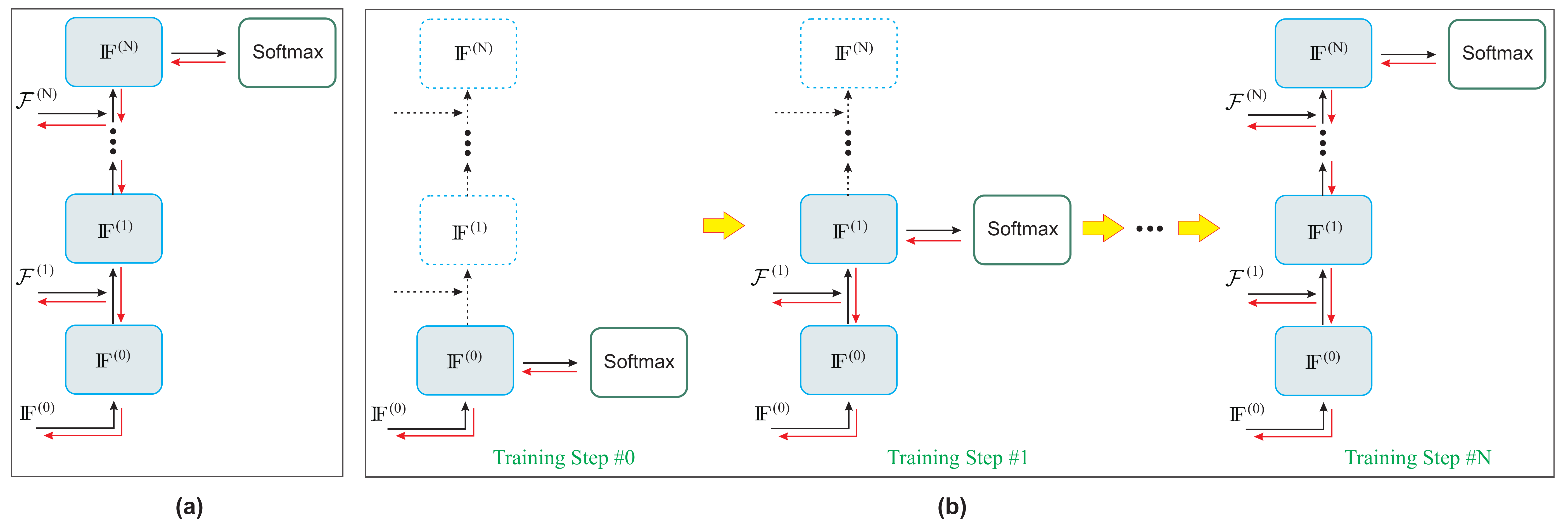}}
	\caption{Illustration of the two network training methods.
(a) Directly train the whole network.
(b) Stepwise optimize the network parameters.
In this figure, the global context memory cell ${\rm I\!F}^{(n)}$ is unfolded over the attention iterations.
The training step $\#n$ corresponds to the $n$-th attention iteration.
The black and red arrows denote the forward and backward passes, respectively.
Some passes, such as those between the two ST-LSTM layers, are omitted for clarity.
Better viewed in colour.
}
	\label{fig:trainmethod}
\end{figure*}

In this part, we first briefly describe the basic training method which directly optimizes the parameters of the whole network,
we then propose a more advanced stepwise training scheme for our GCA-LSTM network.

\subsubsection{Directly Train the Whole Network}
Since the classification is performed by using the last refined global context,
to train such a network,
it is natural and intuitive to feed the action label as the training output at the last attention iteration,
and back-propagate the errors from the last step, i.e., directly optimize the whole network as shown in \figurename{~\ref{fig:trainmethod}(a)}.

\subsubsection{Stepwise Training}
Owing to the recurrent attention mechanism,
there are frequent mutual interactions among different modules (the two ST-LSTM layers and the global context memory cell, see \figurename{~\ref{fig:AttentionGateSTLSTM}}) in our network.
Moreover, during the progress of multiple attention iterations, new parameters are also introduced.
Due to these facts, it is rather difficult to simply optimize all parameters and all attention iterations of the whole network directly as mentioned above.

Therefore, we propose a stepwise training scheme for our GCA-LSTM network, which optimizes the model parameters incrementally.
The details of this scheme are depicted in \figurename{~\ref{fig:trainmethod}(b) and Algorithm 1.

\begin{algorithm}
\caption{Stepwise train the GCA-LSTM network.}
\label{array-sum}
\begin{algorithmic}[1]
    \State Randomly initialize the parameters of the whole network with zero-mean Gaussian.
%
%
%
%
    \For {$n$ = $0$ to $N$} ~~// $n$ is the training step
        \State Feed the action label as the training output at the
        \NoNumber{attention iteration $n$}.
        \Do
        \State Training an epoch: optimizing the parameters used
        \NoNumber{in the iterations $0$ to $n$ via back-propagation}.
        \DoWhile {Validation error is decreasing}
	\EndFor
\end{algorithmic}
\end{algorithm}

The proposed stepwise training scheme is effective and efficient in optimizing the parameters and ensuring the convergence of the GCA-LSTM network.
Specifically, at each training step $n$,
we only need to optimize a subset of parameters and modules which are used by the attention iterations $0$ to $n$.
\footnote{Note that \#0 is not an attention iteration, but the process of initializing the global context memory cell (${\rm I\!F}^{(0)}$).
To facilitate the explantation of the stepwise training, we here temporally describe it as an attention iteration.}
Training this shrunken network is more effective and efficient than directly training the whole network.
At the step $n+1$, a larger scale network needs to be optimized.
However, the training at step $n+1$ is also very efficient,
as most of the parameters and passes have already been optimized (pre-trained well) by its previous training steps.

\section{Two-stream GCA-LSTM Network}
\label{sec:approach2stream}

\begin{figure*}
	\centerline{\includegraphics[scale=0.342]{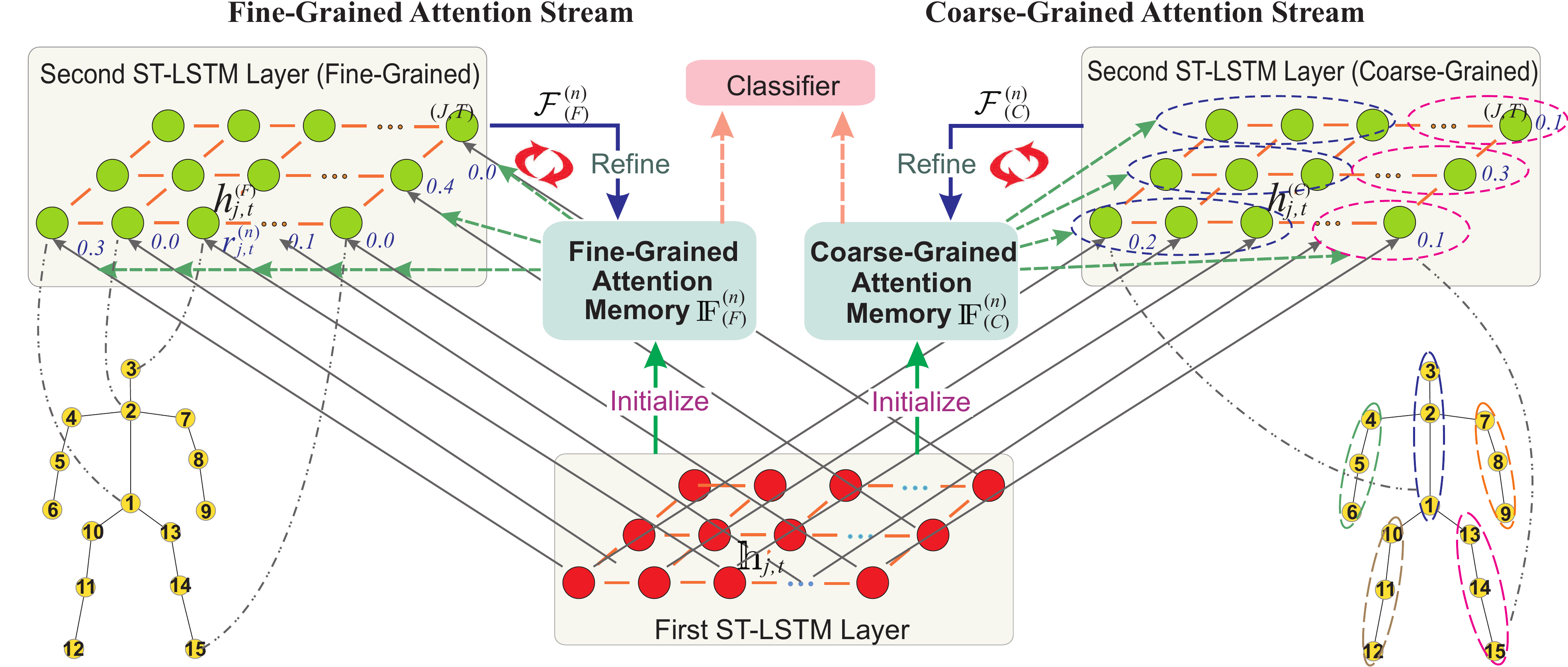}}
	\caption{
Illustration of the two-stream GCA-LSTM network, which incorporates fine-grained (joint-level) attention and coarse-grained (body part-level) attention.
To perform coarse-grained attention, the joints in a skeleton are divided into five body parts,
and all the joints from the same body part share a same informative score.
In the second ST-LSTM layer for coarse-grained attention, we only show two body parts at each frame, and other body parts are omitted for clarity.
}
	\label{fig:AttentionGateSTLSTM_2stream}
\end{figure*}


In the aforementioned design (Section \ref{sec:approach}),
the GCA-LSTM network performs action recognition by selectively focusing on the informative joints in each frame,
i.e., the attention is carried out at joint level (fine-grained attention).
Beside fine-grained attention,
coarse-grained attention can also contribute to action analysis.
This is because some actions are often performed at body part level.
For these actions, all the joints from the same informative body part tend to have similar importance degrees.
For example, the postures and motions of all the joints (elbow, wrist, palm, and finger) from the right hand
are all important for recognizing the action \emph{salute} in the NTU RGB+D dataset \cite{Shahroudy_2016_CVPR},
i.e., we need to identify the informative body part ``right hand'' here.
This implies coarse-grained (body part-level) attention is also useful for action recognition.

As suggested by Du \etal \cite{du2015hierarchical},
the human skeleton can be divided into five body parts (torso, left hand, right hand, left leg, and right leg) based on the human physical structure.
These five parts are illustrated as the right part of \figurename{~\ref{fig:AttentionGateSTLSTM_2stream}}.
Therefore, we can measure the informativeness degree of each body part with regarding to the action sequence, and then perform coarse-grained attention.

Specifically,
we extend the design of out GCA-LSTM model,
and introduce a two-stream GCA-LSTM network here,
which jointly takes advantage of a fine-grained (joint-level) attention stream and a coarse-grained (body part-level) attention stream.

The architecture of the two-stream GCA-LSTM is illustrated in \figurename{~\ref{fig:AttentionGateSTLSTM_2stream}}.
In each attention stream, there is a global context memory cell to maintain the global attention representation of the action sequence,
and also a second ST-LSTM layer to perform attention.
This indicates we have two separated global context memory cells in the whole architecture,
which are respectively the fine-grained attention memory cell (${\rm I\!F}_{(F)}^{(n)}$) and the coarse-grained attention memory cell (${\rm I\!F}_{(C)}^{(n)}$).
The first ST-LSTM layer, which is used to encode the skeleton sequence and initialize the global context memory cells, is shared by the two attention streams.

The process flow (including initialization, attention, and refinement) in the fine-grained attention stream is the same as the GCA-LSTM model introduced in Section \ref{sec:approach}.
The operation in the coarse-grained attention stream is also similar.
The main difference is that,
in the second layer, the coarse-grained attention stream performs attention by selectively focusing on the informative body parts in each frame.

Concretely, in the attention iteration $n$, the network learns an informativeness score ($r_{P,t}^{(n)}$) for each body part $P$ ($P\in\{1,2,3,4,5\}$) as:

\begin{equation}
\label{eq:e_p_t}
e_{P,t}^{(n)} = W_{e_3}
    \left(
        \tanh
        \left(
            W_{e_4}
            \left(
                \begin{array}{ccc}
                \bar{\mathbbm{h}}_{P,t} \\
                {\rm I\!F}_{(C)}^{(n-1)} \\
                \end{array}
            \right)
        \right)
   \right)
\end{equation}
\begin{equation}
\label{eq:r_j_t}
r_{P,t}^{(n)} = \frac{\exp ( e_{P,t}^{(n)} )}{ \sum\limits_{u=1}^5 \sum\limits_{v=1}^T \exp ( {e_{u,v}^{(n)}} ) }
\end{equation}
where $\bar{\mathbbm{h}}_{P,t}$ is the representation of the body part $P$ at frame $t$,
which is calculated based on the hidden representations of all the joints that belong to $P$, with average pooling as:
\begin{equation}
\label{eq:e_p_t_avg}
\bar{\mathbbm{h}}_{P,t} = \frac{1}{J_P}\sum\limits_{j \in P} \mathbbm{h}_{j,t}
\end{equation}
where $J_P$ denotes the number of joints in body part $P$.

To perform coarse-grained attention,
we allow each joint $j$ in body part $P$ to share the informativeness degree of $P$,
i.e., at frame $t$, all the joints in $P$ use the same informativeness score $r_{P,t}^{(n)}$,
as illustrated in \figurename{~\ref{fig:AttentionGateSTLSTM_2stream}}.
Hence, in the coarse-grained attention stream, if $j \in P$, then the cell state of the second ST-LSTM layer is updated at the spatio-temporal step $(j,t)$ as:
\begin{eqnarray}
\label{eq:gatingstlstm_c1}
\nonumber
c_{j, t}  &=&         r_{P, t}^{(n)}  \odot i_{j, t} \odot u_{j, t} \\
          && + ~ (1 - r_{P, t}^{(n)}) \odot f_{j, t}^{(S)} \odot  c_{j-1, t} \\\nonumber
          && + ~ (1 - r_{P, t}^{(n)}) \odot f_{j, t}^{(T)} \odot  c_{j, t-1}
\end{eqnarray}

Multiple attention iterations are also performed in the proposed two-stream GCA-LSTM network.
Finally, the refined fine-grained attention memory ${\rm I\!F}^{(N)}_{(F)}$ and coarse-grained attention memory ${\rm I\!F}^{(N)}_{(C)}$ are both fed to the softmax classifier,
and the prediction scores of these two streams are averaged for action recognition.

The proposed step-wise training scheme can also be applied to this two-stream GCA-LSTM network,
and at the training step $\#n$, we simultaneously optimize the two attention streams, both of which correspond to the $n$-th attention iteration.


\section{Experiments}
\label{sec:experiments}

We evaluate our proposed method on the
NTU RGB+D \cite{Shahroudy_2016_CVPR},
SYSU-3D \cite{jianfang_CVPR15},
UT-Kinect \cite{xia2012view},
SBU-Kinect Interaction \cite{yun2012two},
and Berkeley MHAD \cite{ofli2013berkeley} datasets.
To investigate the effectiveness of our approach, we conduct extensive experiments with the following different network structures:

\begin{itemize}
 \item ``ST-LSTM + Global (1)''.
  This network architecture is similar to the original two-layer ST-LSTM network in \cite{liu2016eccv},
  but the hidden representations at all spatio-temporal steps of the second layer are concatenated and fed to a one-layer feed-forward network to generate a global representation of the skeleton sequence,
  and the classification is performed on the global representation;
  while in \cite{liu2016eccv}, the classification is performed on single hidden representation at each spatio-temporal step (local representation).

   \item ``ST-LSTM + Global (2)''.
  This network structure is similar to the above ``ST-LSTM + Global (1)'',
  except that the global representation is obtained by averaging the hidden representations of all spatio-temporal steps.


 \item ``GCA-LSTM''.
  This is the proposed Global Context-Aware Attention LSTM network.
  Two attention iterations are performed by this network.
  The classification is performed on the last refined global context memory cell.
  The two training methods (\emph{direct training} and \emph{stepwise training}) described in Section \ref{sec:approach:training} are also evaluated for this network structure.

\end{itemize}


In addition, we also adopt the large scale NTU RGB+D and the challenging SYSU-3D as two major benchmark datasets to evaluate the proposed ``two-stream GCA-LSTM'' network.

We use Torch7 framework \cite{collobert2011torch7} to perform our experiments.
Stochastic gradient descent (SGD) algorithm is adopted to train our end-to-end network.
We set the learning rate, decay rate, and momentum to $1.5$$\times$$10^{-3}$, $0.95$, and $0.9$, respectively.
The applied dropout probability \cite{srivastava2014dropout} in our network is set to $0.5$.
The dimensions of the global context memory representation and the cell state of ST-LSTM are both $128$.

\subsection{Experiments on the NTU RGB+D Dataset}
\label{sec:experiments:ntu}

The NTU RGB+D dataset \cite{Shahroudy_2016_CVPR} was collected with Kinect (V2).
It contains more than 56 thousand video samples.
A total of 60 action classes were performed by 40 different subjects.
To the best of our knowledge, this is the largest publicly available dataset for RGB+D based human action recognition.
The large variations in subjects and viewpoints make this dataset quite challenging.
There are two standard evaluation protocols for this dataset:
(1) Cross subject (CS): 20 subjects are used for training, and the remaining subjects are used for testing;
(2) Cross view (CV): two camera views are used for training, and one camera view is used for testing.
To extensively evaluate the proposed method, both protocols are tested in our experiment.

We compare the proposed GCA-LSTM network with state-of-the-art approaches, as shown in \tablename{~\ref{table:resultNTU}}.
We can observe that our proposed GCA-LSTM model outperforms the other skeleton-based methods. 
Specifically, our GCA-LSTM network outperforms the original ST-LSTM network in \cite{liu2016eccv}
by 6.9\% with the cross subject protocol, and 6.3\% with the cross view protocol.
This demonstrates that the attention mechanism in our network brings significant performance improvement.

Both ``ST-LSTM + Global (1)'' and ``ST-LSTM + Global (2)'' perform classification on the global representations,
thus they achieve slightly better performance than the original ST-LSTM \cite{liu2016eccv} which performs classification on local representations.
We also observe ``ST-LSTM + Global (1)'' and ``ST-LSTM + Global (2)'' perform similarly.

The results in \tablename{~\ref{table:resultNTU}} also show that
using the \emph{stepwise training} method can improve the performance of our network in contrast to using the \emph{direct training} method.

\begin{table}[h]
\caption{Experimental results on the NTU RGB+D dataset.}
\label{table:resultNTU}
\centering
\begin{tabular}{|l|c|c|}
\hline
Method & CS & CV  \\
\hline\hline
Skeletal Quads \cite{skeletalQuads} & 38.6\%  & 41.4\% \\
Lie Group \cite{vemulapalli2014liegroup}  & 50.1\% &  52.8\% \\
Dynamic Skeletons \cite{jianfang_CVPR15}  &  60.2\% & 65.2\%  \\
HBRNN \cite{du2015hierarchical}  & 59.1\% & 64.0\% \\
Deep RNN \cite{Shahroudy_2016_CVPR} &  56.3\%  &  64.1\%  \\
Deep LSTM \cite{Shahroudy_2016_CVPR} &   60.7\% & 67.3\%  \\
Part-aware LSTM \cite{Shahroudy_2016_CVPR} & 	62.9\% &	70.3\%  \\
JTM CNN \cite{wang2016action} & 73.4\% & 75.2\% \\
STA Model \cite{song2017end} & 73.4\% & 81.2\% \\
SkeletonNet \cite{ke2017skeletonnet} & 75.9\% & 81.2\%\\
Visualization CNN \cite{liu2017enhanced} &76.0\% & 82.6\% \\
\hline
ST-LSTM \cite{liu2016eccv} & 69.2\%   & 77.7\%  \\
\hline
ST-LSTM + Global (1)  &	 70.5\%	&  79.5\% \\
ST-LSTM + Global (2) & 	 70.7\% &	 79.4\% \\
GCA-LSTM (\emph{direct training}) &  74.3\%	&   82.8\% \\
GCA-LSTM (\emph{stepwise training}) & \textbf{76.1\%}	&  \textbf{84.0\%} \\
\hline
\end{tabular}
\end{table}

We also evaluate the performance of the two-stream GCA-LSTM network, and report the results in \tablename{~\ref{table:resultNTU2stream}}.
The results show that by incorporating fine-grained attention and coarse-grained attention,
the proposed two-stream GCA-LSTM network achieves better performance than the GCA-LSTM with fine-grained attention only.
We also observe the performance of two-stream GCA-LSTM can be improved with the \emph{stepwise training} method.
\begin{table}[h]
\caption{Performance of the two-stream GCA-LSTM network on the NTU RGB+D dataset.}
\label{table:resultNTU2stream}
\centering
\begin{tabular}{|l|c|c|}
\hline
Method & CS & CV  \\
\hline \hline
GCA-LSTM (coarse-grained only)&	  74.1\%	&   81.6\% \\
GCA-LSTM (fine-grained only)&	  74.3\%	&   82.8\% \\
Two-stream GCA-LSTM &	  76.2\%	&   84.7\% \\
Two-stream GCA-LSTM with \emph{stepwise training} &	  77.1\%	&   85.1\% \\
\hline
\end{tabular}
\end{table}

\subsection{Experiments on the SYSU-3D Dataset}
\label{sec:experiments:sysu}

The SYSU-3D dataset \cite{jianfang_CVPR15}, which contains 480 skeleton sequences, was collected with Kinect.
This dataset includes 12 action classes which were performed by 40 subjects.
The SYSU-3D dataset is very challenging as the motion patterns are quite similar among different action classes,
and there are lots of viewpoint variations in this dataset.

We follow the standard cross-validation protocol in \cite{jianfang_CVPR15} on this dataset,
in which 20 subjects are adopted for training the network, and the remaining subjects are kept for testing.
We report the experimental results in \tablename{~\ref{table:resultSYSU}}.
We can observe that our GCA-LSTM network surpasses the state-of-the-art skeleton-based methods in \cite{hu2016ECCV,jianfang_CVPR15,liu2016eccv},
which demonstrates the effectiveness of our approach in handling the task of action recognition in skeleton sequences.
The results also show that our proposed \emph{stepwise training} scheme is useful for our network.

\begin{table}[h]
		\caption{Experimental results on the SYSU-3D dataset.}
		\label{table:resultSYSU}
		\centering
		\begin{tabular}{|l|c|}
			\hline
			Method & Accuracy  \\
			\hline \hline
            LAFF (SKL) \cite{hu2016ECCV}  &  54.2\%  \\
            Dynamic Skeletons \cite{jianfang_CVPR15}  &  75.5\%  \\
			\hline
			ST-LSTM \cite{liu2017skeleton}  &  76.5\%  \\
			\hline
            ST-LSTM + Global (1)   &  76.8\% \\
            ST-LSTM + Global (2)  &	 76.6\% \\
            GCA-LSTM (\emph{direct training}) &  77.8\% \\ 
            GCA-LSTM (\emph{stepwise training}) & \textbf{78.6\%} \\
			\hline
		\end{tabular}
\end{table}

Using this challenging dataset, we also evaluate the performance of the two-stream attention model.
The results in \tablename{~\ref{table:resultSYSU2stream}} show that the two-stream GCA-LSTM network is effective for action recognition.

\begin{table}[h]
\caption{Performance of the two-stream GCA-LSTM network on the SYSU-3D dataset.}
\label{table:resultSYSU2stream}
\centering
\begin{tabular}{|l|c|}
\hline
Method & Accuracy  \\
\hline \hline
GCA-LSTM (coarse-grained only) &   76.9\% \\
GCA-LSTM (fine-grained only) &   77.8\% \\
Two-stream GCA-LSTM  	&   78.8\% \\
Two-stream GCA-LSTM with \emph{stepwise training}  	&   79.1\% \\
\hline
\end{tabular}
\end{table}

\subsection{Experiments on the UT-Kinect Dataset}
\label{sec:experiments:utkinect}

The UT-Kinect dataset \cite{xia2012view} was recorded with a stationary Kinect.
The skeleton sequences in this dataset are quite noisy.
A total of 10 action classes were performed by 10 subjects,
and each action was performed by the same subject twice.

We follow the standard leave-one-out-cross-validation protocol in \cite{xia2012view} to evaluate our method on this dataset.
Our approach yields state-of-the-art performance on this dataset, as shown in \tablename{~\ref{table:resultUTKinectprotocol1}}.

\begin{table}[!htp]
		\caption{Experimental results on the UT-Kinect dataset.}
		\label{table:resultUTKinectprotocol1}
		\centering
		\begin{tabular}{|l|c|}
			\hline
			Method & Accuracy  \\
			\hline\hline
            Grassmann Manifold \cite{slama2015accurate} & 88.5\% \\
			Histogram of 3D Joints \cite{xia2012view} & 90.9\% \\
            Riemannian Manifold \cite{devanne20153} & 91.5\% \\
            Key-Pose-Motifs Mining \cite{Wang_2016_CVPR_Mining} & 93.5\%  \\
            Action-Snippets and Activated Simplices \cite{wang2016recognizing} & 96.5\% \\
            \hline
            ST-LSTM \cite{liu2016eccv} & 97.0\% \\
            \hline
			ST-LSTM + Global (1)   &  97.0\% \\
            ST-LSTM + Global (2)  &	 97.5\% \\
            GCA-LSTM (\emph{direct training})  &  98.5\% \\    
            GCA-LSTM (\emph{stepwise training})  &  \textbf{99.0\%} \\
			\hline
		\end{tabular}
\end{table}

\begin{table*}[!htb]
\caption{Evaluation of robustness against the input noise.
Gaussian noise $\mathcal{N}(0, \sigma^2)$ is added to the 3D coordinates of the skeletal joints.
}
\label{table:addnoiseMHAD}
\centering
\begin{tabular}{|c|c|c|c|c|c|c|c|c|} 
\hline
Standard deviation ($\sigma$) of noise & ~~~$0.1cm$~~  & ~~~$1 cm$~~~  &  ~~~$2 cm$~~~  &  ~~~$4 cm$~~~  &  ~~~$8 cm$~~~  &  ~~~$12 cm$~~~  & ~~~$16 cm$~~~ & ~~~$32 cm$~~~ \\
\hline
\hline
Accuracy & 100\%  & 99.3\%  & 98.5\%  & 97.5\%  & 95.6\%  & 92.7\%  & 80.4\%  & 61.5\%  \\
\hline
\end{tabular}
\end{table*}

\begin{table*}[htb]
\caption{Performance comparison of different attention iteration numbers $(N)$.
}
\label{table:resultAllDSIterNo}
\centering
\begin{tabular}{|c|c|c|c|c|c|} 
\hline
~\#Attention Iteration~ & ~NTU RGB+D (CS)~ &  ~NTU RGB+D (CV)~  &  ~~~~~~UT-Kinect~~~~~~ & ~~~~~~SYSU-3D~~~~~~ & ~~Berkeley MHAD~~ \\
\hline \hline
1  &    72.9\%  &   81.8\%  &   98.0\%  &   77.8\% &   \textbf{100\%}  \\ \hline
2  &    \textbf{76.1\%}  &   \textbf{84.0\%}  &  \textbf{99.0\%}  &   \textbf{78.6\%} &   \textbf{100\%} \\
\hline
\end{tabular}
\end{table*}

\begin{table*}[htb]
\caption{Performance comparison of different parameter sharing schemes.
}
\label{table:resultNTUParaShare}
\centering
\begin{tabular}{|c||c|c||c|c|} 
\hline
\multirow{3}{*}{~\#Attention Iteration~}    & (a) &(b)&(c)&(d)\\
                                            &  w/o sharing within iteration & w/ sharing within iteration  & w/o sharing within iteration &  w/~ sharing within iteration   \\
                                            &  w/~ sharing cross iterations &  w/ sharing cross iterations & w/o sharing cross iterations &  w/o sharing cross iterations   \\
\hline \hline
1  &   71.0\%                       &    72.9\%                       &    \underline{71.0\%}           &72.9\%                                    \\ \hline
2  &    \underline{73.0\%}           &   \underline{74.3\%}                  &    \underline{\textbf{73.4\%}}  &   \underline{\textbf{76.1\%}}              \\ \hline
3  &   \underline{\textbf{73.1\%}}  &   \underline{\textbf{74.4\%}}     &    69.3\%                       &   \underline{73.2\%}                         \\
\hline
\end{tabular}
\end{table*}

\subsection{Experiments on the SBU-Kinect Interaction Dataset}
\label{sec:experiments:cum}

\begin{table}[!bp]
\caption{Experimental results on the SBU-Kinect Interaction dataset.}
\label{table:resultSBU}
\centering
\begin{tabular}{|l|c|}
			\hline
			Method & Accuracy   \\
			\hline \hline
			Yun \etal \cite{yun2012two} & 80.3\% \\
            CHARM \cite{li2015category} & 83.9\% \\
			Ji \etal \cite{ji2014interactive} & 86.9\% \\
			HBRNN \cite{du2015hierarchical}  & 80.4\% \\
            Deep LSTM \cite{zhu2016co} & 86.0\% \\
			Co-occurrence LSTM \cite{zhu2016co} & 90.4\% \\
            SkeletonNet \cite{ke2017skeletonnet} & 93.5\% \\
            \hline
            ST-LSTM \cite{liu2016eccv} & 93.3\% \\
			\hline
            GCA-LSTM (\emph{direct training}) &  94.1\% \\
            GCA-LSTM (\emph{stepwise training}) & \textbf{94.9\%} \\
			\hline
\end{tabular}
\end{table}
The SBU-Kinect Interaction dataset \cite{yun2012two} includes 8 action classes for two-person interaction recognition.
This dataset contains 282 sequences corresponding to 6822 frames.
The SBU-Kinect Interaction dataset is challenging because of (1) the relatively low accuracies of the coordinates of skeletal joints recorded by Kinect,
and (2) complicated interactions between two persons in many action sequences.

We perform 5-fold cross-validation evaluation on this dataset by following the standard protocol in \cite{yun2012two}.
The experimental results are depicted in \tablename{~\ref{table:resultSBU}}.
In this table, HBRNN \cite{du2015hierarchical}, Deep LSTM \cite{zhu2016co}, Co-occurrence LSTM \cite{zhu2016co}, and ST-LSTM \cite{liu2016eccv}
are all LSTM based models for action recognition in skeleton sequences, and are very relevant to our network.
We can see that the proposed GCA-LSTM network achieves the best performance among all of these methods.

\subsection{Experiments on the Berkeley MHAD Dataset}
\label{sec:experiments:mhad}

The Berkeley MHAD dataset was recorded by using a motion capture system.
It contains 659 sequences and 11 action classes, which were performed by 12 different subjects.

We adopt the standard experimental protocol on this dataset, in which 7 subjects are used for training and the remaining 5 subjects are held out for testing.
The results in \tablename{~\ref{table:resultMHAD}} show that our method achieves very high accuracy ($100\%$) on this dataset.

\begin{table}[h]
		\caption{Experimental results on the Berkeley MHAD dataset}
		\label{table:resultMHAD}
		\centering
		\begin{tabular}{|l|c|c|}
			\hline
			Method & Accuracy  \\
			\hline \hline
			Ofli \etal \cite{ofli2014sequence}   & 95.4\% \\
            Vantigodi \etal \cite{vantigodi2013real}   & 96.1\%  \\
			Vantigodi \etal \cite{vantigodi2014action}    & 97.6\%  \\
			Kapsouras \etal \cite{kapsouras2014action}   & 98.2\%  \\
            \hline
			ST-LSTM \cite{liu2016eccv}  & 100\%  \\
			\hline
            GCA-LSTM (\emph{direct training}) &   100\% \\
            GCA-LSTM (\emph{stepwise training}) & 100\% \\
			\hline
		\end{tabular}
\end{table}
As the Berkeley MHAD dataset was collected with a motion capture system rather than a Kinect,
thus the coordinates of the skeletal joints are relatively accurate.
To evaluate the robustness with regarding to the input noise,
we also investigate the performance of our GCA-LSTM network on this dataset by adding zero mean input noise to the skeleton sequences,
and show the results in \tablename{~\ref{table:addnoiseMHAD}.
We can see that even if we add noise with the standard deviation ($\sigma$) set to $12 cm$ (which is significant noise in the scale of human body),
the accuracy of our method is still very high ($92.7\%$).
This demonstrates that our method is quite robust against the input noise.

\subsection{Evaluation of Attention Iteration Numbers}
\label{sec:experiments:attentioniternum}

We also test the effect of different attention iteration numbers on our GCA-LSTM network,
and show the results in \tablename{~\ref{table:resultAllDSIterNo}}.
We can observe that increasing the iteration number can help to strength the classification performance of our network
(using 2 iterations obtains higher accuracies compared to using only 1 iteration).
This demonstrates that the recurrent attention mechanism proposed by us is useful for the GCA-LSTM network.

Specifically, we also evaluate the performance of 3 attention iterations by using the large scale NTU RGB+D dataset,
and the results are shown in \tablename{~\ref{table:resultNTUParaShare}}.
We find the performance of 3 attention iterations is slightly better than 2 iterations if we share the parameters over different attention iterations
(see columns (a) and (b) in \tablename{~\ref{table:resultNTUParaShare}}).
This consistently shows using multiple attention iterations can improve the performance of our network progressively.
We do not try more iterations due to the GPU's memory limitation.

We also find that if we do not share the parameters over different attention iterations (see columns (c) and (d) in \tablename{~\ref{table:resultNTUParaShare}}),
then too many iterations can bring performance degradation (the performance of using 3 iterations is worse than that of using 2 iterations).
In our experiment, we observe the performance degradation is caused by over-fitting (increasing iteration number will introduce new parameters if we do not share parameters).
But the performance of two iterations is still significantly better than one iteration in this case.
We will also give the experimental analysis of the parameter sharing schemes detailed in Section \ref{sec:experiments:paramsharing}.





\subsection{Evaluation of Parameter Sharing Schemes}
\label{sec:experiments:paramsharing}

As formulated in Eq. (\ref{eq:e_j_t}),
the model parameters $W_{e_1}$ and $W_{e_2}$ are introduced for calculating the informativeness score at each spatio-temporal step in the second layer.
Also multiple attention iterations are carried out in this layer.
To regularize the parameter number inside our network and improve the generalization capability,
we investigate two parameter sharing strategies for our network:
(1) Sharing within iteration: $W_{e_1}$ and $W_{e_2}$ are shared by all spatio-temporal steps in the same attention iteration;
(2) Sharing cross iterations: $W_{e_1}$ and $W_{e_2}$ are shared over different attention iterations.
%
We investigate the effect of these two parameter sharing strategies on our GCA-LSTM network,
and report the results in \tablename{~\ref{table:resultNTUParaShare}}.

In \tablename{~\ref{table:resultNTUParaShare}}, we can observe that:
(1) Sharing parameters within iteration is useful for enhancing the generalization capability of our network,
  as the performance in columns (b) and (d) of \tablename{~\ref{table:resultNTUParaShare}} is better than (a) and (c), respectively.
(2) Sharing parameters over different iterations is also helpful for handling the over-fitting issues,
but it may limit the representation capacity,
as the network with two attention iterations which shares parameters within iteration but does not share parameters over iterations achieves the best result
(see column (d) of \tablename{~\ref{table:resultNTUParaShare}}).
As a result, in our GCA-LSTM network, we only share the parameters within iteration, and two attention iterations are used.


\subsection{Evaluation of Training Methods}
\label{sec:experiments:trainmethods}

The previous experiments showed that
using the \emph{stepwise training} method can improve the performance of our network in contrast to using \emph{direct training}
(see \tablename{~\ref{table:resultNTU}, \ref{table:resultUTKinectprotocol1}, \ref{table:resultSYSU}, \ref{table:resultSBU}}).
To further investigate the performance of these two training methods,
we plot the convergence curves of our GCA-LSTM network in \figurename{~\ref{fig:losscurve}}.

We analyze the convergence curves (\figurename{~\ref{fig:losscurve}}) of the \emph{stepwise training} method as follows.
By using the proposed \emph{stepwise training} method,
at the training step $\#0$,
we only need to train the subnetwork for initializing the global context (${\rm I\!F}^{(0)}$),
i.e., only a subset of parameters and modules need to be optimized,
thus the training is very efficient and the loss curve converges very fast.
When the validation loss stops decreasing, we start the next training step $\#1$.
Step $\#1$ contains new parameters and modules for the first attention iteration, which have not been optimized yet, therefore, loss increases immediately at this epoch.
However, most of the parameters involved at this step have already been pre-trained well by the previous step $\#0$,
thus the network training is quite effective,
and the loss drops to a very low value after only one training epoch.

By comparing the convergence curves of the two training methods,
we can find (1) the network converges much faster if we use \emph{stepwise training}, compared to \emph{directly train} the whole network.
We can also observe that (2) the network is easier to get over-fitted by using \emph{direct training} method,
as the gap between the train loss and validation loss starts to rise after the $20$th epoch.
These observations demonstrate that the proposed \emph{stepwise training} scheme is quite useful for effectively and efficiently training our GCA-LSTM network.

\begin{figure}
	\centerline{\includegraphics[scale=0.58]{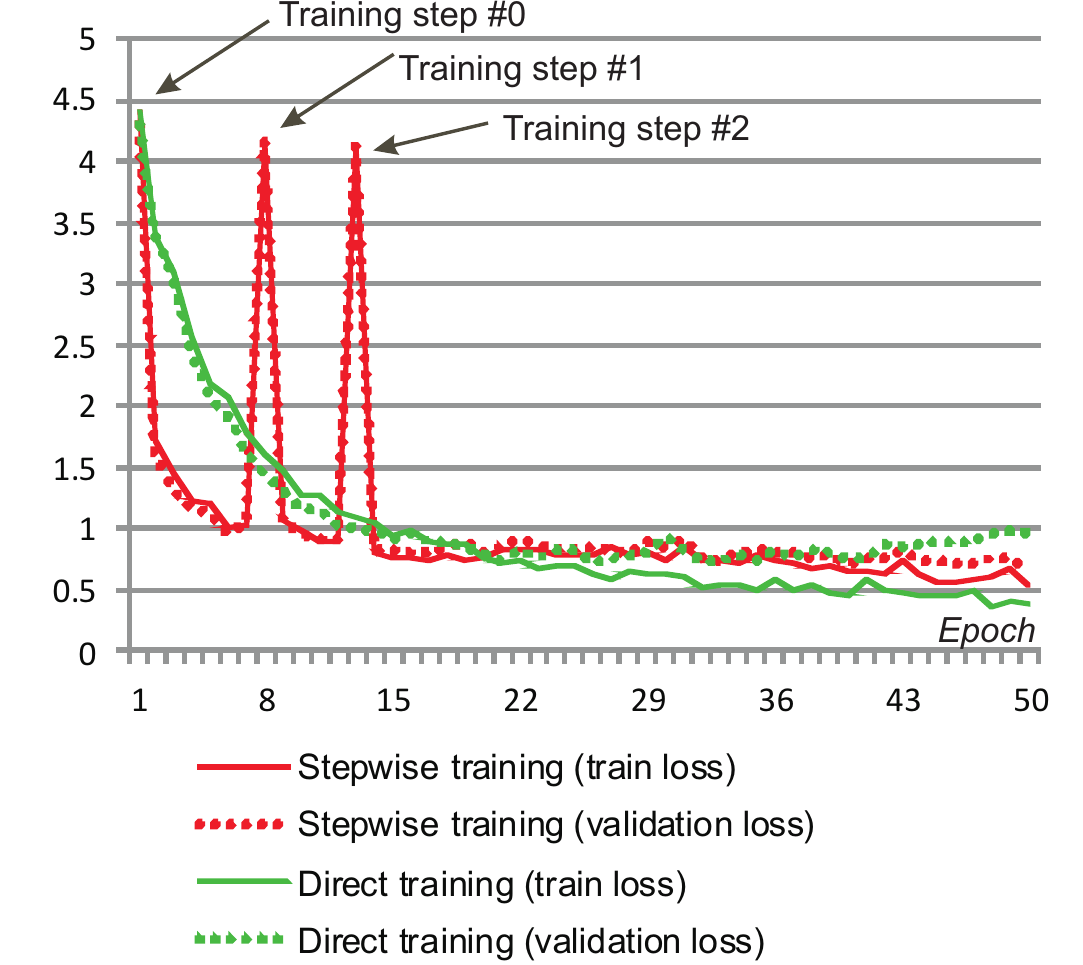}}
	\caption{Convergence curves of the GCA-LSTM network with two attention iterations by respectively using \emph{stepwise training} (in red) and \emph{direct training} (in green) on the NTU RGB+D dataset.
Better viewed in colour.
}
	\label{fig:losscurve}
\end{figure}

\subsection{Evaluation of Initialization Methods and Attention Designs}
\label{sec:experiments:trainmethods}

In Section \ref{sec:approach:init},
we introduce two methods to initialize the global context memory cell (${\rm I\!F}^{(0)}$).
The first is averaging the hidden representations of the first layer (see Eq. (\ref{eq:initialglobalcontext})),
and the second is using a one-layer feed-forward network to obtain ${\rm I\!F}^{(0)}$.
We compare these two initialization methods in \tablename{~\ref{table:resultDiffMethInitialize}}.
The results show that these two methods perform similarly.
In our experiment, we also find that by using feed-forward network, the model converges faster,
thus the scheme of feed-forward network is used to initialize the global context memory cell in our GCA-LSTM network.  

\begin{table}[h]
		\caption{Performance comparison of different methods of initializing the global context memory cell.}
		\label{table:resultDiffMethInitialize}
		\centering
		\begin{tabular}{|l|c|c|c|}
			\hline
			Method & NTU RGB+D (CS) & NTU RGB+D (CV) \\
			\hline \hline
            Averaging &   73.8\%     &    83.1\%   \\
            Feed-forward network & 74.3\%     &    82.8\%   \\
			\hline
		\end{tabular}
\end{table}

In the GCA-LSTM network, the informativeness score $r_{j,t}^{(n)}$ is used as a gate within LSTM neuron, as formulated in Eq. ({\ref{eq:gatingstlstm_c}}).
We also explore to replace this scheme with soft attention method \cite{xu2015show,luong2015effective},
i.e., the attention representation $\mathcal{F}^{(n)}$ is calculated as $\sum_{j=1}^J\sum_{t=1}^T r_{j,t}^{(n)} \mathbbm{h}_{j,t}$.
Using soft attention, the accuracy drops about one percentage point on the NTU RGB+D dataset.
This can be explained as equipping LSTM neuron with gate $r_{j,t}^{(n)}$ provides LSTM better insight about when to update, forget or remember.
In addition, it can keep the sequential ordering information of the inputs $\mathbbm{h}_{j,t}$,
while soft attention loses ordering and positional information.

\subsection{Visualizations}
\label{sec:experiments:discussions}

\begin{figure*}
	\centerline{\includegraphics[scale=0.16]{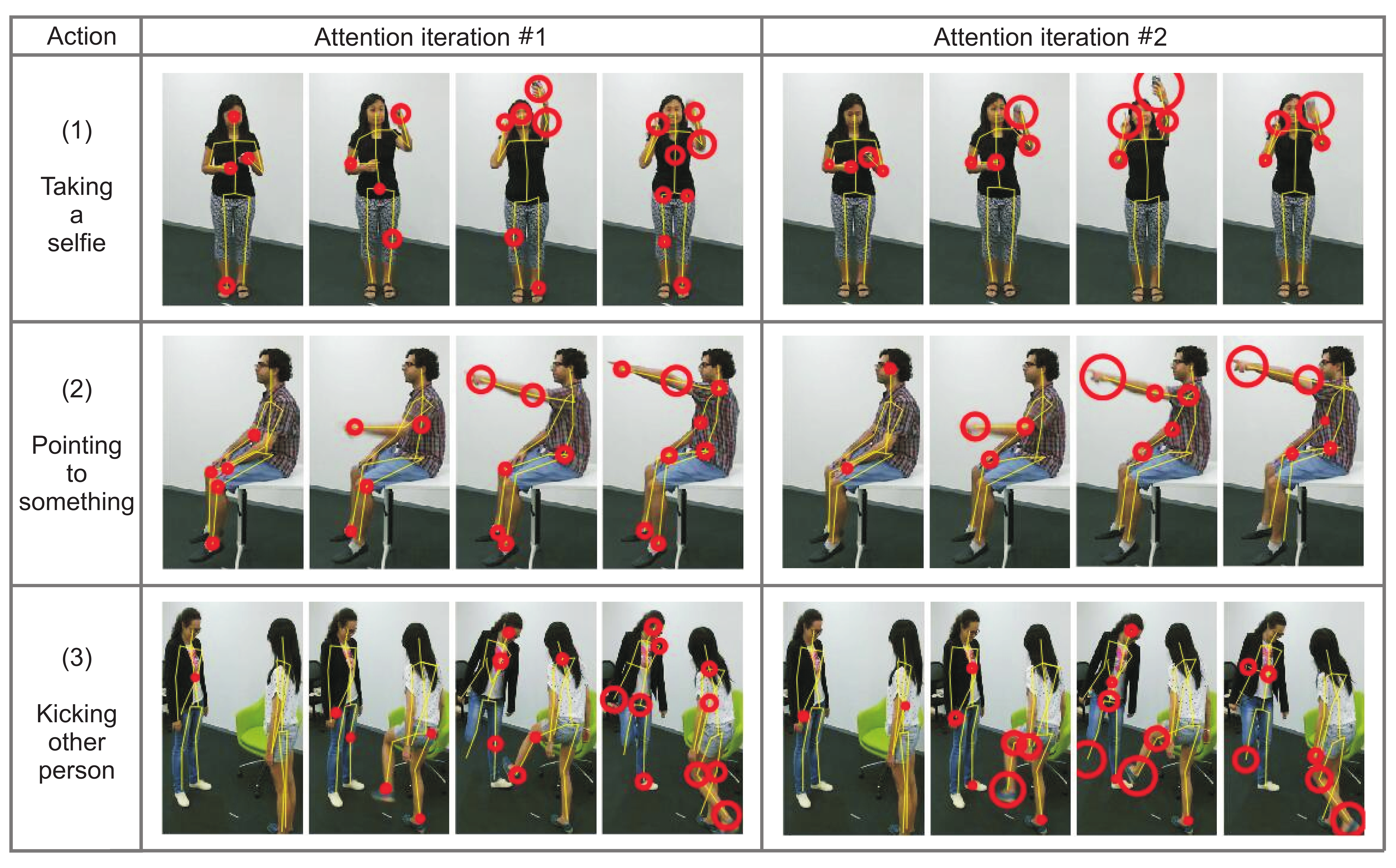}}
	\caption{Examples of qualitative results on the NTU RGB+D dataset.
Three actions (\emph{taking a selfie}, \emph{pointing to something}, and \emph{kicking other person}) are illustrated.
The informativeness scores of two attention iterations are visualized.
Four frames are shown for each iteration.
The circle size indicates the magnitude of the informativeness score for the corresponding joint in a frame.
For clarity, the joints with tiny informativeness scores are not shown.
}
	\label{fig:dominatejoint_details}
\end{figure*}

To better understand our network,
we analyze and visualize the informativeness score evaluated by using the global context information on the large scale NTU RGB+D dataset in this section.

We analyze the variations of the informativeness scores over the two attention iterations to verify the effectiveness of the recurrent attention mechanism in our method,
and show the qualitative results of three actions (\emph{taking a selfie}, \emph{pointing to something}, and \emph{kicking other person}) in \figurename{~\ref{fig:dominatejoint_details}}.
The informativeness scores are computed with soft attention for visualization.  
In this figure, we can see that the attention performance increases between the two attention iterations.
In the first iteration, the network tries to identify the potential informative joints over the frames.
After this attention, the network achieves a good understanding of the global action.
Then in the second iteration, the network can more accurately focus on the informative joints in each frame of the skeleton sequence.
We can also find that the informativeness score of the same joint can vary in different frames.
This indicates that \emph{our network performs attention not only in spatial domain, but also in temporal domain}.

In order to further quantitatively evaluate the effectiveness of the attention mechanism,
we analyze the classification accuracies of the three action classes in \figurename{~\ref{fig:dominatejoint_details}} among all the actions.
We observe if the attention mechanism is not used, the accuracies of these three classes are 67.7\%, 71.7\%, and 81.5\%, respectively.
However, if we use one attention iteration, the accuracies rise to 67.8\%, 72.4\%, and 83.4\%, respectively.
If two attention iterations are performed, the accuracies become 67.9\%, 73.6\%, and 86.6\%, respectively.

\begin{figure}
	\centerline{\includegraphics[scale=1.1]{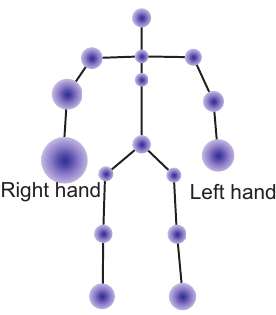}}
	\caption{Visualization of the average informativeness gates for all testing samples.
The size of the circle around each joint indicates the magnitude of the corresponding informativeness score.}
	\label{fig:dominatejoint}
\end{figure}

To roughly explore which joints are more informative for the activities in the NTU RGB+D dataset,
we also average the informativeness scores of the same joint in all the testing sequences,
and visualize it in \figurename{ \ref{fig:dominatejoint}}.
We can observe that averagely, more attention is assigned to the hand and foot joints.
This is because in the NTU RGB+D dataset, most of the actions are related to the hand and foot postures and motions.
We can also find that the average informativeness score of the right hand joint is higher than that of left hand joint.
This indicates most of the subjects are right-handed.


\section{Conclusion}
\label{sec:conclusion}
In this paper, we have extended the original LSTM network to construct a Global Context-Aware Attention LSTM (GCA-LSTM) network for skeleton based action recognition,
which has strong ability in selectively focusing on the informative joints in each frame of the skeleton sequence with the assistance of global context information.
Furthermore, we have proposed a recurrent attention mechanism for our GCA-LSTM network, in which the selectively focusing capability is improved iteratively.
In addition, a two-stream attention framework is also introduced.
The experimental results validate the contributions of our approach by achieving state-of-the-art performance on five challenging datasets.

\section*{Acknowledgement}
\label{sec:acknowledgement}
This work was carried out at the Rapid-Rich Object Search (ROSE) Lab at Nanyang Technological University (NTU), Singapore.
The ROSE Lab is supported by the National Research Foundation, Singapore,
under its Interactive Digital Media (IDM) Strategic Research Programme.
We acknowledge the support of NVIDIA AI Technology Centre (NVAITC) for the donation of the Tesla K40 and K80 GPUs used for our research at the ROSE Lab.
Jun Liu would like to thank Qiuhong Ke from University of Western Australia for helpful discussions.

\ifCLASSOPTIONcaptionsoff
  \newpage
\fi



%

\bibliographystyle{IEEEtran}
\end{document}